\documentclass{ecai}
\usepackage{algorithm2e}
\usepackage{times}
\usepackage{graphicx}
\usepackage{latexsym}
\graphicspath{{../pdf/}{../jpeg/}}
\DeclareGraphicsExtensions{.pdf}
\usepackage[cmex10]{amsmath}
\usepackage{array}
\usepackage{bm}
\usepackage{color, colortbl}
\usepackage{diagbox}
\usepackage{mdwmath}
\usepackage{mdwtab}
\usepackage{eqparbox}
\usepackage{url}
\usepackage{amssymb,amsfonts,amsthm,mathrsfs}
\usepackage{booktabs}
\usepackage{multirow}
\usepackage{tikz}
\usetikzlibrary{bayesnet}
\usepackage{tkz-graph}

\ecaisubmission  

\newcommand{\B}[1]{\boldsymbol #1}

\begin{document}

\title{Relational Neural Machines\thanks{In proceedings of ECAI 2020.}}

\author{Giuseppe Marra\institute{Department of Information Engineering, University of Florence, Florence, email: g.marra@unifi.it} \and Michelangelo Diligenti\institute{Department of Information Engineering and Science, University of Siena, Siena, email: \{diligmic,fgiannini,marco,maggini\}@diism.unisi.it} \and Francesco Giannini\footnotemark[3] \and\\  Marco Gori\footnotemark[3] \and Marco Maggini\footnotemark[3]}

\maketitle
\bibliographystyle{ecai}

\begin{abstract}
Deep learning has been shown to achieve impressive results in several tasks where a large amount of training data is available. However, deep learning solely focuses on the accuracy of the predictions, neglecting the reasoning process leading to a decision, which is a major issue in life-critical applications.
Probabilistic logic reasoning allows to exploit both statistical regularities and specific domain expertise to perform reasoning under uncertainty, but its scalability and brittle integration with the layers processing the sensory data have greatly limited its applications.
For these reasons, combining deep architectures and probabilistic logic reasoning is a fundamental goal towards the development of intelligent agents operating in complex environments.
This paper presents Relational Neural Machines, a novel framework allowing to jointly train the parameters of the learners and of a First--Order Logic based reasoner. A Relational Neural Machine is able to recover both classical learning from supervised data in case of pure sub-symbolic learning, and Markov Logic Networks in case of pure symbolic reasoning, while allowing to jointly train and perform inference in hybrid learning tasks. Proper algorithmic solutions are devised to make learning and inference tractable in large-scale problems. The experiments show promising results in different relational tasks.
\end{abstract}

\section{Introduction}
\label{sec:introduction}
In the last few years, the availability of a large amount of supervised data caused a significant improvement in the performances of sub-symbolic approaches like artificial neural networks. In particular, deep neural networks have  achieved impressive results in several tasks, thanks to their ability to jointly learn the decision function and the data representation from the low-level perception inputs~\cite{goodfellow2016deep,lecun2015deep}. However, the dependency on the amount and quality of training data is also a major limitation of this class of approaches.
Standard neural networks can struggle to represent relational knowledge on different input patterns, or relevant output structures, which have been shown to bring significant benefits in many challenging applications like image segmentation tasks~\cite{chen2015learning}. For this reason, several work has been done in the direction of learning and representing relations using embeddings~\cite{jameel2016entity,minervini2017regularizing,wang2015knowledge,dumanvcic2017demystifying,nickel2016holographic,allamanis2017learning} and in developing and injecting relational features into the learning process~\cite{santoro2017simple, niepert2016discriminative}.

On the other hand, symbolic approaches~\cite{DeRaedtProbLog2007,kimmig2012short,muggleton1994inductive} are generally based on probabilistic logic reasoners, and can express high-level relational dependencies in a certain domain of discourse and perform exact or approximate inference in presence of uncertainty.
Markov Logic Networks (MLN)~\cite{richardson2006markov} and its variants like Probabilistic Soft Logic~\cite{bach2017hinge} are relational undirected models, mapping First--Order Logic formulas to a Markov network, and allowing to train the parameters of the reasoner and perform inference under uncertainty.

Another related line of research studies hybrid approaches leveraging neural networks to learn the structure of the reasoning process like done, for instance, by Relational Restricted Boltzmann machines~\cite{kaur2017relational} and auto-encoding logic programs~\cite{dumanvcic2019learning}. Similarly, Neural Markov Logic Networks~\cite{marra2019neural} extend MLN by defining the potential functions as general neural networks which are trained together with the model parameters.
Neural Theorem Prover~\cite{rocktaschel2016learning, rocktaschel2017end} is an end-to-end differentiable prover that shows state-of-the-art performances on some link prediction benchmarks by combining Prolog backward chain with a soft unification scheme. 
TensorLog~\cite{yang2017differentiable,kathryn2018tensorlog} is a recent framework to reuse the deep learning infrastructure of TensorFlow to perform probabilistic logical reasoning.

Whereas the previously discussed methods provide a large step forward in the definition of a flexible and data-driven reasoning process, they do not still allow to co-optimize the low-level learners processing the environmental data.
Methods bridging the gap between symbolic and sub-symbolic levels are commonly referred as neuro-symbolic approaches~\cite{garcez2012neural, kaur2017relational, sourek2018lifted}.
An early attempt to integrate learning and reasoning is the work by Lippi et al.~\cite{lippi2009betaresidue}. The main limitation of this work is that it was devised ad-hoc to solve a specific task in bioinformatics and it does not define a general methodology to apply it to other contexts.

A methodology to inject logic into deep reinforcement learning has been proposed by Jiang et al. \cite{jiang2019neural}, while a distillation method to inject logic knowledge into the network weights is proposed by Hu et al.~\cite{hu2016harnessing}. Deep Structured Models~\cite{chen2015learning} define a schema to inject complex output structures into deep learners. The approach is general but it does not focus on logic knowledge but on imposing statistical structure on the output predictions. Hazan et al.~\cite{hazan2016blending} integrate learning and inference in Conditional Random Fields~\cite{sutton2012introduction}, but they also do not focus on logic reasoning.
The Semantic Loss~\cite{xu2017semantic} allows to translate the desired output structure of a learner via the definition of a loss, which can also accommodate logic knowledge. However, the loss and the resulting reasoning process is fixed, thus limiting the flexibility of the approach.
Deep ProbLog~\cite{manhaeve2018deepproblog} is a neuro-symbolic approach, based on the probabilistic logic programming language ProbLog~\cite{DeRaedtProbLog2007} and approximating the predicates via deep learning. This approach is very flexible but it is limited to cases where exact inference is possible, as it lacks a modular and scalable solution like the one proposed in this paper.

Deep Logic Models (DLM)~\cite{marra2019integrating} are instead capable of jointly training the sensory and reasoning layers in a single differentiable architecture, which is a major advantage with respect to related approaches like Semantic-based Regularization~\cite{diligenti2017semantic}, Logic Tensor Networks~\cite{donadello2017logic} or Neural Logic Machines~\cite{dong2018neural}.
However, DLM is based on a brittle stacking of the learning and reasoning modules, failing to provide a real tight integration on how low-level learner employs the supervised data. For this reason, DLM requires the employment of heuristics like training plans to make learning effective.

This paper presents Relational Neural Machines (RNM), a novel framework introducing fundamental improvements over previous state-of-the-art-models in terms of scalability and in the tightness of the connection between the trainer and the reasoner.
A RNM is able to perfectly replicate the effectiveness of training from supervised data of standard deep architectures, while still co-training a reasoning module over the environment that is built during the learning process. The bonding is very general as any (deep) learner can be integrated and any output or input structure can be expressed.
On the other hand, when restricted to pure symbolic reasoning, RNM can replicate the expressivity of Markov Logic Networks~\cite{richardson2006markov}. 

The outline of the paper is as follows. Section~\ref{sec:dlm} presents the model and how it can be used to integrate logic and learning.
Section~\ref{sec:training_and_inference} studies tractable approaches to perform inference and model training from supervised and unsupervised data. 
Section~\ref{sec:exp_results} shows the experimental evaluation of the proposed ideas on various datasets.
Finally, Section~\ref{sec:conclusions} draws some conclusions and highlights some planned future work.

\section{Model}
\label{sec:dlm}
A Relational Neural Machine establishes a probability distribution over a set of $n$ output variables of interest $\B y = \{ y_1, \ldots, y_n \}$, given a set of predictions made by one or multiple deep architectures, and the model parameters.
In this paper the output variables are assumed to be binary, i.e. $y_i = \{0,1\}$, but the model can be extended to deal with continuous values for regression tasks.

Unlike standard neural networks which compute the output via a simple forward pass, the output computation in an RNM can be decomposed into two stages: a \emph{low-level} stage processing the input patterns, and a subsequent \emph{semantic} stage, expressing constraints over the output and performing higher level reasoning.
In this paper, it is assumed that there is a single network processing the input sensorial data, but the theory is trivially extended to any number of learners.
The first stage processes $D$ input patterns $\B x = \{\B x_1, \ldots, \B x_D\}$, returning the values $\B f$ using the network with parameters $\B w$. The higher layer takes as input $\B f$ and applies reasoning using a set of constraints, whose parameters are indicated as $\B \lambda$, then it returns the set of output variables $\B y$.

A RNM model defines a conditional probability distribution in the exponential family defined as:
\begin{equation}
p(\B y | \B f, \B \lambda) =
\frac{1}{Z} \exp\left(\sum_{c} \lambda_c \Phi_c(\B f, \B y) \right)
\label{eq:dlm}
\end{equation}
where $Z$ is the partition function and the $C$ potentials $\Phi_c$ express some properties on the input and output variables. The parameters $\B \lambda = \{\lambda_1, \ldots, \lambda_C\}$ determine the strength of the potentials $\Phi_c$.

This model can express a vast range of typical learning tasks. We start reviewing how to express simple classification problems, before moving to general neuro-symbolic integration mixing learning and reasoning. A main advantage of RNMs is that they can jointly express and solve these use cases, which are typically been studied as stacked separate problems.

In a classical and pure supervised learning setup, the patterns are i.i.d., it is therefore possible to split the $\B y, \B f$ into disjoint sets grouping the variables of each pattern, forming separate cliques. Let us indicate as $\B y(x), \B f(x)$ the portion of the output and function variables referring to the processing of an input pattern $x$.
A single potential $\Phi_0$ is needed to represent supervised learning, and this potential decomposes over the patterns as:
\begin{equation}
    \Phi_0(\B y, \B f) = \sum_{x \in S} \phi_0(\B y(x), \B f(x)) \ .
    \label{eq:supervised_potential}
\end{equation}
where $S \subseteq \B x$ is the set of supervised patterns. This yields the distribution,
\begin{equation}
p_0(\B y | \B f, \B \lambda) = \frac{1}{Z} \exp\left(\displaystyle\sum_{x \in S} \phi_0(\B y(x), \B f(x))\right)
\label{eq:classification_model}
\end{equation}

\paragraph{One-label classification. }
The mutual exclusivity rule requires to assign a zero probability to assignments stating that a pattern can belong to more than one class. The following potential is defined for any generic input pattern $x$:
\[
\phi_0(\B y(x), \B f(x)) \!=\! \left\{
\begin{array}{lr}
\!\!-\infty &\!\!\mbox{ if } \exists i,j \!:\! y_i(x)\!=\!y_j(x)\!=\!1, i\!\neq\! j \\
\!\!\B f(x) \!\cdot\! \B y & \mbox{otherwise}
\end{array}
\right.
\]
When only the $\Phi_0$ potential is used, each pattern corresponds to a set of outputs independent on the other pattern outputs given the $\B f$, the partition function decomposes over the patterns and the probability distribution $p_0$ simplifies to:
\[
p_0(\B y | \B f, \B \lambda) =  \left\{
\begin{array}{lr}
0 & \hspace{-4cm}\mbox{if } \exists x,i,j : y_i(x)\!=\!y_j(x)\!=\!1, i\neq j \\
\frac{\exp\left(\displaystyle\sum_{x \in S} \B f(x)\cdot\B y(x)\right)}{\displaystyle\sum_{y}\prod_{x \in S} \exp\left(\B f(x)\cdot\B y(x)\right)} = & \!\!\mbox{otherwise}\\
= \displaystyle\prod_{x \in S} \frac{\exp\left(\B f(x)\cdot\B y(x)\right)}{\displaystyle\sum_{i \in Y}  \exp\left(f^i_w(x)\cdot y_i(x)\right)} =&\\
=\displaystyle\prod_{x \in S} softmax(\B f(x), y(x)) &
\end{array}
\right.
\]
This result provides an elegant justification for the usage of the softmax output for networks used in one-label classification tasks.

\paragraph{Multi-label.}
The following potential is expressed for each input pattern:
$\phi_0(\B y(x), \B f(x)) = \B f(x) \cdot \B y(x)$.

When plugging in the previously defined potential into the potential in Equation~\ref{eq:supervised_potential} and the result plugged into Equation~\ref{eq:classification_model}, the partition function can be decomposed into one component for each pattern and class, since each pattern and classification output is independent on all the other classifications:
\[
	\begin{array}{lcl}
	p_0(\B y | \B f, \B \lambda) &\!\!=\!\!&
	\frac{\exp\left(\displaystyle\sum_{x \in S} \B f(x)\cdot \B y(x) \right)}{\exp\left( \displaystyle\sum_{x \in S} \displaystyle\sum_{\B y(x)} \B f(x) \cdot \B y(x)\right)}
	=\\
	&=& \displaystyle\prod_{x \in S} \displaystyle\prod_{i \in Y}
	\frac{\exp\left( f_i(x) \cdot y_i(x) \right)}{1 + \exp\left(f_i(x)\right)} =\\
	&=&\displaystyle\prod_{x \in S} \Big[\displaystyle\prod_{i \in Y^+(x)} \!\!\!\!\sigma(f_i(x)) \cdot\!\!\!\! \displaystyle\prod_{i \in Y^-(x)} \!\!\!\!\left(1 - \sigma(f_i(x))\right)\Big]
	\end{array}
\]
where $\sigma$ is the sigmoid function, $Y^+(x),Y^-(x)$ are the set of positive and negative classes for pattern $x$. This result provides an elegant justification for the usage of a sigmoidal output layer for multi-label classification tasks.

\subsection{Neuro-symbolic integration}
\label{sec:neuro}
The most interesting and general case is when the presented model is used to perform both learning and reasoning, which is a task referred in the literature as neuro-symbolic integration.
\begin{table}
	\[
	\begin{array}{|l|}
	\hline
	\mbox{Image Correlations} \forall O \in {\boldmath \mathcal{O}} \\
	\forall i_1 \forall i_2 ~ SameLoc(i_1, i_2) \Rightarrow \big( O(i_1) \land O(i_2) \big) \lor \big( \lnot O(i_1) \land \lnot O(i_2) \big)  \\
	\hline
	\mbox{Knowledge Graphs}\\
	\forall i ~O(i) \Rightarrow Mammal(i)  ~~~~ \forall O \in \{ Lion, Cat, Dog, Sheep, \ldots \}\\
	\forall i ~Mammal(i) \Rightarrow Animal(i)\\
	\forall i ~Mammal(i) \Rightarrow Legs(i) \land Body(i) \land Tail(i) \\
	\hline
	\mbox{General Knowledge}\\
	\forall i ~Lion(i) \rightarrow Savanna(i) \lor Zoo(i) \\
	\forall i ~Wall(i) \rightarrow \lnot Savanna(i) \\
	\hline
	\mbox{Category Correlations}\\
	\forall i ~PolarBear(i) \land Lion(i)\\
	\forall i ~Antelope(i) \land Lion(i)\\
	\hline
	\mbox{Supervisions}\\
	Lion(i1), Wall(i1), \ldots \\
	\hline
	\end{array}
	\]
	\label{tab:neuro_symbolic_integration_example}
	\caption{An example of the knowledge available to express an object detection task. where $\boldmath \mathcal{O}$ is the set of all possible objects (or predicates to be learned). Supervisions about the image $i1$ containing the objects $Lion$, and $Wall$ is added together to the background knowledge.}
\end{table}

The general model described in Equation~\ref{eq:dlm} is materialized with one potential $\Phi_0$ enforcing the consistency with the supervised data together with potentials representing the logic knowledge.
Using a similar approach to Markov Logic Networks, a set of First--Order Logic (FOL) formulas is input to the system, and  there is a potential $\Phi_c$ for each formula.
The general form of the conditional probability distribution becomes:
\begin{equation}
p(\B y | \B f, \B \lambda) = \frac{1}{Z} \exp\left(\displaystyle\sum_{x \in S} \phi_0(\B f(x), \B y(x)) + \sum_{c} \lambda_c \Phi_c(\B y) \right)
\label{eq:dlm_neuro_symbolic_integration}
\end{equation}
where it is assumed that some (or all) the predicates in a KB are unknown and need to be learned together with the $\B \lambda$ parameters driving the reasoning process.

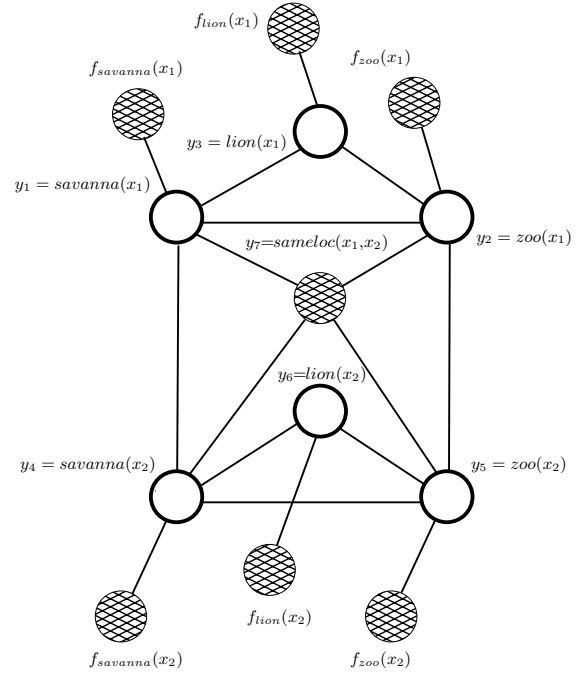
\begin{figure}[t!]
\centering
    \scalebox{0.7}[0.7]{
        \begin{tikzpicture}[scale=1.0]
        {
            \node at (0,0) {\includegraphics[width=0.4\textwidth]{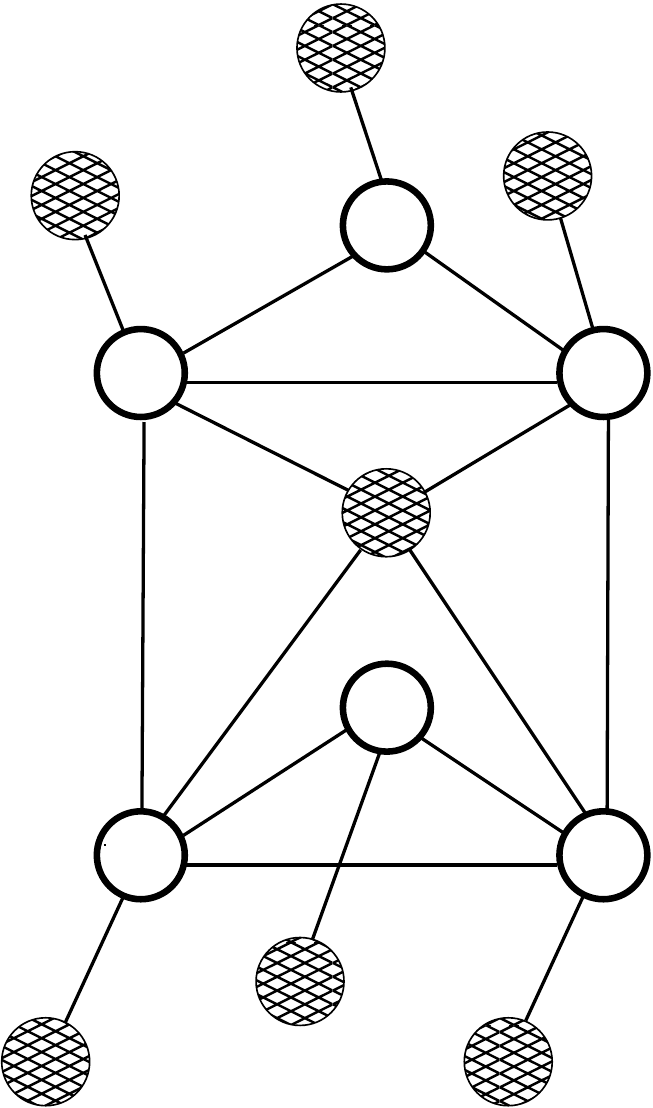}};
            \node at (-2.8,+4.8) (n1) {$f_{savanna}(x_1)$} ;
            \node at (+1.8,+5.0) (n2) {$f_{zoo}(x_1)$} ;
            \node at (-1.1,+5.7) (n3) {$f_{lion}(x_1)$} ;
            \node at (-2.8,-6.4) (n4) {$f_{savanna}(x_2)$} ;
            \node at (+1.8,-6.4) (n5) {$f_{zoo}(x_2)$} ;
            \node at (-0.1,-5.6) (n6) {$f_{lion}(x_2)$} ;
            \node at (+0.6,+1.5) (n13)  {$y_7\!\!=\!\!sameloc(x_1,\!x_2)$} ;
            \node at (-3.8,+2.6) (n7)  {$y_1=savanna(x_1)$} ;
            \node at (+4.5,+1.6) (n8)  {$y_2=zoo(x_1)$} ;
            \node at (-0.9,+3.4) (n9)  {$y_3=lion(x_1)$} ;
            \node at (-3.7,-2.7) (n10) {$y_4=savanna(x_2)$} ;
            \node at (+4.4,-2.7) (n11) {$y_5=zoo(x_2)$} ;
            \node at (+0.7,-1.0) (n12) {$y_6\!\!=\!\!lion(x_2)$} ;
        }
        \end{tikzpicture}
    }
    \caption{The graphical model representing a RNM, where the output variables $\B y$ depend on the output of first stage $\B f$, processing the inputs $\{x_1,x_2\}$ instantiated for the rules $\forall x ~ lion(x) \Rightarrow savanna(x) \lor wall(x)$, $\forall x \forall x^\prime ~ sameloc(x,x^\prime) \land savanna(x) \Rightarrow savanna(x^\prime)$, and $\forall x \forall x^\prime ~ sameloc(x,x^\prime) \land zoo(x) \Rightarrow zoo(x^\prime)$.}
    \label{fig:dlm_graphical_model}
\end{figure}
A \emph{grounded} expression (the same applies to atom or predicate) is a FOL rule whose variables are assigned to specific constants.
It is assumed that the undirected graphical model has the following structure, each grounded atom corresponds to a node in the graph, and all nodes connected by at least one rule are connected on the graph, so that there is one clique (and then potential) for each grounding $g_c$ of the formula in $\B y$. It is assumed that all the potentials resulting from the $c$-th formula share the same weight $\lambda_c$, 
therefore the potential $\Phi_c$ is the sum over all groundings of $\phi_c$ in the world $\B y$, such that: $\Phi_c(\B y) = \sum_{\B y_{c,g}} \phi_c(\B y_{c.g})$ where $\phi_c(g_c)$ assumes a value equal to $1$ and $0$ if the grounded formula holds true and false. This yields the probability distribution:
\[
p(\B y | \B f, \B \lambda) \!=\! \frac{1}{Z} \exp\!\left(\!\displaystyle\sum_{x \in S} \phi_0(\B f(x), \B y(x)) \!+\! \sum_{c} \lambda_c \!\displaystyle\sum_{\B y_{c,g}} \phi_c(\B y_{c,g}) \!\right)
\]

\paragraph{Example.} It is required to train a classifier detecting the objects on images for a multi-object detection task in real world pictures. A knowledge graph may be available to describe hierarchical dependencies among the object classes, or object compositions. Pictures may be correlated by the locations where they have been shot.
Table~\ref{tab:neuro_symbolic_integration_example} shows the knowledge that could be used to express such a task, where the unknown predicates to be trained are indicated as the set $\boldmath \mathcal{O}$. Other predicates like $SameLoc$ may be known a priori based the meta-information attached to the images.
Figure~\ref{fig:dlm_graphical_model} shows the graphical model correlating the output variables $\B y$ and the $\B f$for the the inputs $\{x_1,x_2\}$ instantiated for the rules $\forall x ~ lion(x) \Rightarrow savanna(x) \lor zoo(x)$, $\forall x \forall x^\prime ~ sameloc(x,x^\prime) \land savanna(x) \Rightarrow savanna(x^\prime)$, and $\forall x \forall x^\prime ~ sameloc(x,x^\prime) \land zoo(x) \Rightarrow zoo(x^\prime)$.
The goal of the training process is to train the classifiers approximating the predicates, but also to establish the relevance of each rule. For example, the formula $\forall x ~Antelope(x) \land Lion(x)$ is likely to be associated to a higher weight than $\forall x ~PolarBear(x) \land Lion(x)$, which are unlikely to correlate in the data. 

\paragraph{Logic Tensor Networks.}
Logic Tensor Networks (LTN)~\cite{serafini2017learning} is a framework to learn neural networks under the constraints imposed by some prior knowledge expressed as a set of FOL clauses. As shown in this paragraph, LTN is a special case of a RNM, when the $\B \lambda$ parameters are frozen. In particular, an LTN expresses each FOL rule via a continuous relaxation of a logic rule using fuzzy logic. The strength $\lambda_c$ of the rule is assumed to be known a priori and not trained by the LTN. These rules provide a prior for the functions. Therefore, assuming the $\B \lambda$ parameters are fixed, an LTN considers the following distribution:
\begin{equation*}
\begin{array}{lcl}
p(\B f | \B y) &\propto& p(\B y | \B f) \cdot p(\B f) = \\
&=&\overbrace{\frac{1}{Z_1} \exp\left(\sum_{x \in S}\phi_0(y(x), f(x))\right)}^{p(\B y | \B f)} \cdot \\
&&\cdot \overbrace{\frac{1}{Z_2} \exp\left(\sum_{c} \lambda_c \Phi^s_c(\B f) \right)}^{p(\B f)}
\end{array}
\label{eq:ltn}
\end{equation*}
where $\Phi^s_c$ is the continuous relaxation of the $c$-th logic rule, $p(\B y | \B f)$ is used to express the fitting of the supervised data and the prior $p(\B f)$ gives preference to the functions respecting the logic constraints. The parameters $\B w$ of the $\B f$ of an LTN can be optimized via gradient ascent by maximizing the likelihood of the training data.

\paragraph{Semantic Based Regularizaion.}
Semantic-Based Regularization (SBR)~\cite{diligenti2017semantic}, defines a learning and reasoning framework which allows to train neural networks under the constraints imposed by the prior logic knowledge. The declarative language
Lyrics~\cite{marra2019lyrics} is available to provide a flexible and easy to use frontend for the SBR framework.
At training time, SBR employs the knowledge like done by LTN, while SBR uses a continuous relaxation $\Phi^s_c$ of the $c$-th logic rule and of the output vector at inference time.
Therefore, SBR can also be seen as a special instance of a RNM, when the $\B \lambda$ parameters are frozen and the continuous relaxation of the logic is used at test time. Both LTN and SBR have a major disadvantage over RNM, as they can not learn the weights of the reasoner, which are required to be known a priori. This is very unlikely to happen in most of the real world scenarios, where the strength of each rule must be co-trained with the learning system.

\section{Learning and Inference}
\label{sec:training_and_inference}
\paragraph{Training.}
A direct model optimization in RNM is intractable in most interesting cases, as a the computation of the partition function requires a summation over all possible assignments of the output variables.
However, if a partition function is assumed to be factorized into separate independent groups of variables $\B y_i \in \B y$, it holds that:
\[
Z \approx \displaystyle\prod_i Z_i
\]

A particularly interesting case is when it is assumed that the partition function factorizes over the potentials like done in piecewise likelihood~\cite{sutton2007piecewise}:
\[
Z \approx \displaystyle\prod_c Z_c = \prod_{c} \left[ \sum_{\B y_c^\prime} \exp(\lambda_c \Phi_c(\B f, \B y_c^\prime)) \right]
\]
where $\B y_c$ is the subset of variables in $\B y$ that are involved in the computation of $\Phi_c$.
We indicate as piecewise-local probability for the $c$-th constraint:
\begin{equation}
p^{PL}_c(\B y_c|\B f, \B \lambda) = \frac{\exp\left( \lambda_c \Phi_c(\B f, \B y_c) \right)}{Z_c} =  \frac{\exp\left( \lambda_c \Phi_c(\B f, \B y_c) \right)}{\displaystyle\sum_{\B y_c^\prime} \exp\left( \lambda_c \Phi_c(\B f, \B y^\prime_c) \right)}
\label{eq:piecewise_local_pro}
\end{equation}

Under this assumption, the factors can be distributed over the potential giving the following generalized piecewise likelihood:
\[\begin{array}{lcl}
p(\B y | \B f, \B \lambda) &\approx& \displaystyle\prod_c p(\B y_c | \B y\backslash \B y_c, \B f, \lambda_c) = \displaystyle\prod_c p^{PL}_c(\B y_c|\B f, \B \lambda) 
\end{array}
\]

If the variables in $\B y$ are binary, the computation of $Z$ requires summation over all possible assignments which has $O(2^{|\B y|})$ complexity.
Using the local decomposition this is reduced to $O(2^n)$, where $n$ is the size of the largest potential.
When a single potential involves too many variables, the pseudo-likelihood decomposition can be used, where each variable is factorized into a separate component with linear complexity with respect to the numbers variables:
\[
p^{PPL}_c(\B y_c|\B f, \B \lambda) = \prod_{y_i \in \B y_c} \frac{\lambda_c \Phi_c(\B f, \B y_c)}{ \displaystyle\sum_{b=\{0,1\}} \exp\left( \lambda_c \Phi_c(\B f, \B [y_i=b,\B y_c \backslash y_i])\right) }
\]
where the factorization is performed with respect of the single variables, which has a cost proportional to $n$.

Assuming that the constraints are divided into two groups $\{C_1, C_2\}$, for which the local piecewise partitioning $PL$ and the pseudo-likelihood $PPL$ approximations are used, the distribution becomes:
\[
p(\B y | \B f, \B \lambda) \approx \prod_{c \in C_1} p^{PL}_c(\B y_c|\B f, \lambda_c) \cdot \prod_{c \in C_2} p^{PPL}_c(\B y_c|\B f, \lambda_c)
\]

If the $c$-th constraint is factorized using the $PL$ partitioning, the derivatives of the log-likelihood with respect to the model potential weights are:
\begin{eqnarray}
\frac{\partial \log p(\B y | \B f, \B \lambda)}{\partial \lambda_c} &\!\!\!\!\approx\!\!\!\!&
\Phi_c(\B f, \B y) - \displaystyle\sum_{\B y^\prime_c \in \B Y_c} p^{PL}_c(\B y^\prime_c |\B f, \B \lambda) \cdot \Phi_c(\B f, \B y^\prime_c) = \nonumber\\
&\!\!\!\!=\!\!\!\!& \Phi_c(\B f, \B y) - E_{p^{PL}_c}[\Phi_c] \label{eq:lambda_derivative}
\end{eqnarray}
and with respect to the learner parameters:
\begin{equation}
\begin{array}{lcl}
\frac{\partial \log p(\B y | \B f, \B \lambda)}{\partial w_k} &\approx&
\displaystyle\sum_i y_i \frac{\partial f_i}{\partial w_k} - E_{p^{PL}_0}\left[ \displaystyle\sum_i y_i \frac{\partial f_i}{\partial w_k} \right] = \\
&=& \displaystyle\sum_i \frac{\partial f_i}{\partial w_k} \left[ y_i - E_{y^\prime \sim p^{PL}_0}\left[ y^\prime_i \right] \right]
\end{array}
\label{eq:w_derivative}
\end{equation}
 
In the following of this section, it is assumed that all potentials are approximated using the piecewise local approximation to keep the notation simple, the extension to the pseudo likelihood is trivial, it is enough to replace the $p_c^{PL}$ with the $p_c^{PLL}$ in Equation~\ref{eq:lambda_derivative}.

\paragraph{Training for neuro-symbolic integration.}
An interesting case is when a potential represents the level of satisfaction of a logical constraint over its groundings in the world $\B y$. In this case the predicates of the $c$-th formula are grounded with a set of $\B x_c$ groundings, and $G_c(\B y)$ indicates the set of outputs for the grounded predicates in the world $\B y$. Therefore, the potential is the sum over the grounded formulas:
\[
\Phi_c(\B y) = \sum_{\B y_{c,g} \in G_c(\B y)} \phi_c(\B y_{c,g}) ,
\]
where $\phi_c(\B y_{c,g}) \in \{0, 1\}$ is the satisfaction of the formula (False or True) by the grounded predicates $\B y_{c,g}$.
\begin{table*}[th!]
        \centering
        \begin{tabular}{c|c|c|c}
                \hline
                \diagbox{\bfseries op}{\bfseries t-norm} &\bfseries Product & \bfseries G\"{o}del & \bfseries \L ukaseiwicz\\ 
                \hline
                $x \land y$ & $x \cdot y$ & $\min(x,y$) & $\max(0, x+y-1$)\\
                \hline
                $x \lor y$ & $x+y-x \cdot y$ & $\max(x, y)$ & $\min(1, x + y)$\\
                \hline
                $\lnot x$ & $1 - x$ & $1 - x$ & $1 - x$\\
                \hline
                $x \Rightarrow y$ & $x\leq y?1:\frac{y}{x}$ & $x\leq y?1:y$ & $\min(1,1-x+y)$\\
                \hline
        \end{tabular}
        \caption{The algebraic operations corresponding to primary logical connectives for the fundamental t-norm fuzzy logics.}
        \label{tab:forward_operations}
\end{table*}
Therefore, each grounding corresponds to a separate potential, even if they are all sharing the same weight. Assuming that each grounding of a formula is independent on all the others, then we can approximate the $Z_c$ as:
\begin{align*}
    Z_c \approx {Z^l_c}^{|G_c(\B y)|} = \left( \displaystyle\sum_{\B y^\prime_{c,g}} \exp(\lambda_c \phi_c(\B y^\prime_{c,g})) \right)^{|G_c(\B y)|}
\end{align*}
where ${|G_c(\B y)|}$ are the total number of groundings of the $c$-th formula in $\B y$ and each grounded formula shares the same local partition function $Z^l_c$. $Z^l_c$ can be efficiently computed by pre-computing $n^+_c,n^-_c$, indicating the number of possible different grounding assignments satisfying or not satisfying the $c$-th formula. Clearly, since for a formula with $n_c$ atoms, there are $2^{n_c}$ possible assignments, it holds that $n^-_c + n^+_c= 2^{n_c}$, yielding:

\begin{equation*}
\begin{array}{lcl}
    Z^l_c &=& \displaystyle\sum_{\B y^\prime_{c,g}} \exp(\lambda_c \phi_c(\B y^\prime_{c,g})) =\\
    &=& \underbrace{n^-_c}_{\substack{\text{each \texttt{False} evaluates} \\  \text{to } e^0 = 1}} + \underbrace{n^+_c e^{\lambda_c}}_{\substack{\text{each \texttt{True} evaluates} \\  \text{to } e^{\lambda_c}}} =
    n^-_c + n^+_c e^{\lambda_c}
\end{array}
\end{equation*}


Using the piecewise local approximation for each grounding, the derivatives with respect of the model parameters become:
\[
\begin{array}{lcl}
\frac{\partial \log p(\B y | \B f, \B \lambda)}{\partial \lambda_c} &\!\!\approx\!\!&
\displaystyle\sum_{\B y_{c,g}} \left[ \phi_c(\B y_{c,g}) - \prod_{\B y^\prime_{c,g}} p^{PL}_{c,g}(\B y^\prime_{c,g}) \phi_c(\B y^\prime_{c,g}) \right] =\\
&\!\!=\!\!& \displaystyle\sum_{\B y_{c,g}} \phi_c(\B y_{c,g}) - |G_c(\B y)| \cdot \mathbb{E}_{p^{PL}_{c,g}}[\phi_c]
\end{array}
\]
Let us indicate as $Avg(\Phi_c, \B y)=\frac{1}{|G_c(\B y)|}\displaystyle\sum_{\B y_{c,g} \in G_c(\B y)} \phi_c(\B y_{c,g})$ the average satisfaction of the $c$-th constraint over the data training data, then the gradient is null when for all constraints:
\begin{equation}
Avg(\Phi_c, \B y) = \mathbb{E}_{p^{PL}_{c,g}}[\phi_c]
\label{eq:dlm_optimal_condition}    
\end{equation}

The expected value of the satisfaction of the formula for a grounding, $\mathbb{E}_{p^{PL}_{c,g}}[\phi_c]$ can be a efficiently computed for a $\{0,1\}$-valued $\phi_c$ as:
\begin{align*}
    \mathbb{E}_{p^{PL}_{c,g}}[\phi_c] &= \displaystyle\sum_{\B y^\prime_{c,g}} p^{PL}_{c,g}(\B y^\prime_{c,g}) \phi_c(\B y^\prime_{c,g}) = \\
	&= \displaystyle\sum_{\B y^\prime_{c,g}} \frac{1}{Z^l_{c}} \exp\left(\lambda_c \phi_c(\B y^\prime_{c,g})\right) \phi_c(\B y^\prime_{c,g}) =\\
	&= \frac{n^+_c e^{\lambda_c}}{Z^l_{c}} =
	\frac{n^+_c e^{\lambda_c}}{n^-_c + n^+_c e^{\lambda_c}}
	\label{eq:dlm_expectation}
\end{align*}
yielding the following optimal assignment to the $c$-th parameter for a given assignment $\B y$:
\begin{equation}
    \lambda_c = \log \frac{Avg(\Phi_c, \B y)}{1 - Avg(\Phi_c, \B y)} - \log \frac{n^+_c}{n^-_c}
    \label{eq:dlm_optimal_parameter}
\end{equation}
which shows that the log-likelihood is maximized by selecting a $\lambda_c$ equal to difference between the log odds of the constraint satisfaction of the data and the log odds of the prior satisfaction of the constraint if all assignments are equally probable.

When the world is fully observed during training, $\B y^T$ indicates the training data assignments, then substituting $\B y = \B y^T$ into Equation~\ref{eq:dlm_optimal_parameter} returns the maximum likelihood assignment for the parameters.

\begin{algorithm}[t]
\KwData{Input data $\B x$, output variables $\B y$, training and observed data $\B y^T$, function models with weights $\B w$}
\KwResult{Trained model parameters $\{\B \lambda, \B w\}$}
Initialize $i=0$, $\B \lambda=\B 0$, random $\B w$;\\
\While{not converged $\land ~ i<max\_iterations$}{
\textbf{Expectation}:\\
~~Compute function outputs $\B f$ on $\B x$ using weights $\B w$; \\
~~Compute MAP solution using Equation~\ref{eq:dlm_map} ${\B y}^\star  = \text{arg}\!\max_{\B y} \log p({\B y} | \B f, \B y^T, \B \lambda)$;\\
\textbf{Maximization}:\\
~~ $\forall c$ compute constraint satisfaction $Avg(\Phi_c, \B y \cup \B y^T$);\\
~~ $\forall c$ compute $\lambda_c$ using Equation~\ref{eq:dlm_optimal_parameter} on $\B y \cup \B y^T$;\\
~~ Backpropagation with respect of the $\B w$ weights using derivatives from Equation~\ref{eq:w_derivative};\\
~~ Update $\B w$;\\
Set i=i+1;
}
\caption{Iterative algorithm to train the function weights $\B w$ and the constraint weights $\B \lambda$.}
\label{al:training}
\end{algorithm}
When the world is not fully observed during training, an iterative EM schema can be used to marginalize over the unobserved data in the expectation step using the inference methodology as described in the next paragraph. Then, the average constraint satisfaction can be recomputed, and then the $\B \lambda, \B w$ parameters can be updated in the maximization step. This process is then iterated until convergence. Algorithm~\ref{al:training} reports the complete training algorithm for RNMs.

\paragraph{Inference.}
The MAP inference process searches the most probable assignment of the $\B y$ given the evidence and the fixed parameters $\B w, \B \lambda$.
The problem of finding the best assignment $\B y^\star$ to the unobserved query variables given the evidence $\B y^e$ and current parameters can be stated as:
\begin{equation}
    \B y^\star =
    arg\displaystyle\max_{\B y^\prime} \sum_{c} \lambda_c \Phi_c(\B f, [\B y^\prime, \B y^e])
    \label{eq:dlm_map}
\end{equation}
where $[\B y^\prime, \B y^e]$ indicates a full assignment to the $\B y$ variables, split into the query and evidence sets.

Gradient-based techniques can not be readily used to optimize the MAP problem stated by Equation~\ref{eq:dlm_map}, since the problem is discrete.
A possible solution could be to relax the $\B y$ values into the $[0,1]$ interval and assume that each potential $\Phi_c(\B f, \B y)$ has a continuous surrogate $\Phi_c^s(\B f, \B y)$ which collapses into the original potential when the $\B y$ assume crisp values and is continuous with respect to each $y_i$. As described in the following, continuous surrogates are very appropriate to describe the potentials representing logic knowledge for neuro-symbolic integration and probabilistic logic reasoning.

When relaxing the potentials to accept continuous input variables, the MAP solution can be found by gradient-based techniques by computing the derivative with respect of each output variable:
\begin{equation}
\frac{\partial \log p(\B y^\prime | \B y^e, \B f, \B \lambda)}{\partial y_k} =
f_k + \displaystyle\sum_c \lambda_c \frac{\partial \Phi^s_c(\B f, [\B y^\prime, \B y^e])}{\partial y_k}
\label{eq:dlm_map_gradient}    
\end{equation}

\paragraph{T-Norms Fuzzy Logics.}

Fuzzy logics extend the set of Boolean logic truth-values $\{0,1\}$ to the unit interval $[0,1]$ and, as a consequence, they can be exploited to convert Boolean logic expressions into continuous and differentiable ones. In particular, a t-norm fuzzy logic~\cite{hajek2013metamathematics} is defined upon the choice of a certain t-norm~\cite{Klement2000}. A t-norm is a binary operation generalizing to continuous values the Boolean logic conjunction ($\wedge$), while it recovers the classical AND when the variables assume the crisp values $0$ (false) or $1$ (true).

Throughout this paper, we assume that given a certain 
variable $a$ assuming a continuous value $a\in[0,1]$, its negation $\neg a$ (also said strong negation) is evaluated as $1-a$. Moreover, a t-norm and the strong negation allows the definition of additional logical connectives. For instance, the implication ($\Rightarrow$) may be defined as the residuum of the t-norm, while the OR ($\vee$) operator, also called t-conorm, may be defined according to the DeMorgan law with respect to the t-norm and the strong negation.
\[
a\Rightarrow b = \sup\{c:a\wedge c\leq b\};\quad a\vee b=\neg(\neg a\wedge\neg b) \ .
\]

Different t-norm fuzzy logics have been proposed in the literature. Table~\ref{tab:forward_operations} reports the operations computed by different logic operators for the three fundamental continuous t-norms, i.e. Product, G\"{o}del and \L ukasiewicz logics.
Furthermore, a fragment of the \L ukasiewicz logic~\cite{giannini2018convex} has been recently proposed for translating logic inference into a differentiable optimization problem, since it defines a large class of clauses which are translated into convex functions.


An important role defining different ways to aggregate logical propositions on (possibly large) sets of domain variables is played by quantifiers.
The universal quantifier $(\forall)$ and the existential quantifier $(\exists)$ express the fact that a clause should hold true over all or at least one grounding. Both the universal and existential quantifier are generally converted into real-valued functions according to different aggregation functions, e.g. the universal one as a t-norm and the existential one as a t-conorm over the groundings.
When multiple universally or existentially quantified variables are present, the conversion is recursively performed from the outer to the inner variables as already stated. For example, consider the rule
\[
\forall x\,A(x)\vee\left(B(x)\wedge\neg C(x)\right)
\]
where $A, B, C$ are three unary predicates defined on the input set $\{x_1,\ldots,x_m\}$. In this case, the output vector $\B y$ is defined as follows,
\[
 [y_A(x_1), \ldots , y_A(x_m), y_B(x_1), \ldots , y_B(x_m), y_C(x_1), \ldots , y_C(x_m) ]
\]
where $y_P(x_i)$ is the output of predicate $P$ when grounded with $x_i$.
The continuous surrogate for the FOL rule grounded over all patterns in the domain, in case of the product t-norm and universal quantifier converted with the arithmetic mean, is given by:
\[
\Phi^s(\B y) = \frac{1}{m} \sum_{i=1}^m y_A(x_i) + (1-y_A(x_i))\cdot y_B(x_i) \cdot (1-y_C(x_i)) \ .
\]


\section{Experiments}
\label{sec:exp_results}

The proposed model has been experimentally evaluated on two different datasets where the relational structure on the output or input data may be exploited. 

\subsection{MNIST Following Pairs}
\label{exp:following}
\begin{figure}[t!]
\centering
    \includegraphics[width=0.45\textwidth]{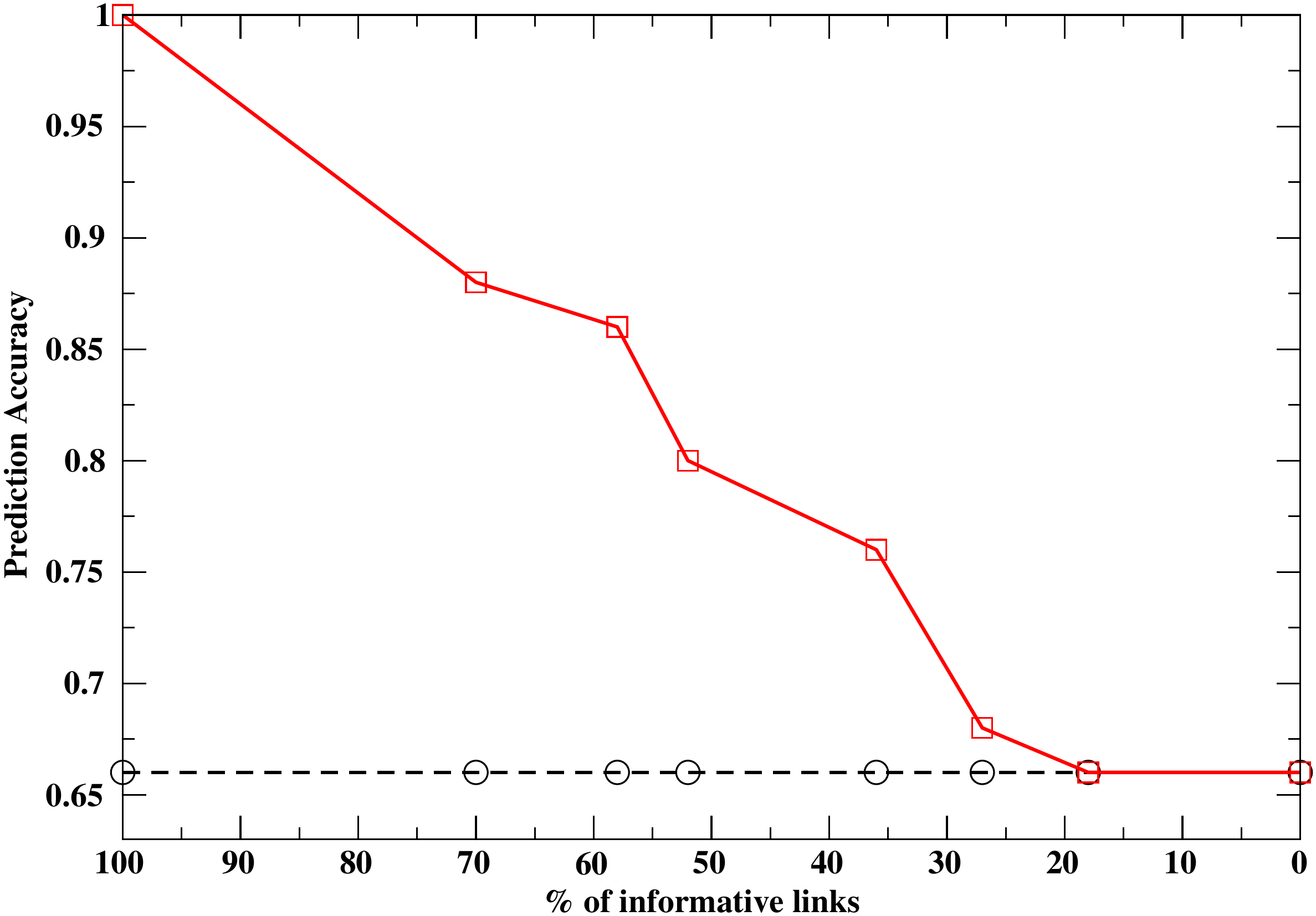}
    \caption{Prediction accuracy with respect to the percentage of links that correctly predict the next digit condition.}
    \label{fig:mnist}
\end{figure}
This small toy task is designed to highlight the capability of RNMs to learn and employ soft rules that are holding only for a sub-portion of the whole dataset. 
The MNIST dataset
contains images of handwritten digits, and this task assumes that additional relational logic knowledge is available to reason over the digits. In particular, given a certain subset of images, a binary predicate $link$ between image pairs is considered. Given two images $x,y$, whose corresponding digits are denoted by $i,j$, a link between $x$ and $y$ is established if the second digit follows the first one, i.e. $i=j+1$.
However, it is assumed that the $link$ predicate is noisy, therefore for $i\neq j+1$, there is a given degree of probability that the $link(x,y)$ is established anyway.
The knowledge about the $link$ predicate can be represented by the following FOL formula
\[
\forall x\forall y\forall i\forall j\,link(x,y)\wedge digit(x,i)\wedge digit(y,j)\Rightarrow i=j+1 \ ,
\]
where $digit(x,i)$ is a binary predicate indicating if a number $i$ is the digit class of the image $x$. 
Since the $link$ predicate holds true also for pairs of non-consecutive digits, the above rule is violated by a certain percentage of digit pairs. Therefore, the manifold established by the $link$ predicate can help in driving the prediction, but the noisy links force the reasoner to be flexible about how to employ the knowledge.

The training set is created by randomly selecting 50 images from the MNIST dataset and by adding the $link$ relation with an incremental degree of noise. 
For each degree of noise in the training set, we created an equally sized test set with the same degree of noise. A neural network with $100$ hidden sigmoid neurons is used to process the input images. 

Figure~\ref{fig:mnist} reports a comparison between RNM and the baseline provided by the neural network varying the percentage of links that are predictive of a digit to follow another one. When the link predicate only holds for consecutive digit pairs, RNM is able to perfectly predict the images on the test set using this information. When the link becomes less informative (more noisy), RNM is still able to employ the rule as a soft suggestion. However, when the percentage of predictive links approaches $10\%$, the relation is not informative at all, as it does not add any information on top of the prior probability that two randomly picked up numbers follow each other. In this case, RNM is still able to detect that the formula is not useful, and only the supervised data is used to learn the predictions. As a result, the predictive accuracy of RNM matches the one of the neural network.


\subsection{Document Classification on the Citeseer dataset. }
\label{exp:CiteSeer}
The CiteSeer dataset
~\cite{lu2003link} 
is a collection of $3312$ scientific papers, each one assigned to one of the $6$ classes: $AG, AI, DB, IR, ML$ and $HCI$. The papers connect to each other by a citation network which contains $4732$ links. Each paper in the dataset is described via its bag-of-words, e.g. a vector where the $i$-th element has a value equal to $1$ or $0$, depending on whether the $i$-th word in the vocabulary is present or not present in the document, respectively. The overall dictionary for this experiment contains $3703$ unique words. 
The domain knowledge used for this task state that connected papers $p_1,p_2$ tend to be about the same topic:
\begin{equation*}
\begin{array}{l}
\forall p_1 \forall p_2 ~AG(p_1) \land Cite(p_1, p_2) \rightarrow AG(p_2)\\
\forall p_1 \forall p_2 ~AI(p_1) \land Cite(p_1, p_2) \rightarrow AI(p_2)\\
\forall p_1 \forall p_2 ~DB(p_1) \land Cite(p_1, p_2) \rightarrow DB(p_2)\\
\forall p_1 \forall p_2 ~IR(p_1) \land Cite(p_1, p_2) \rightarrow IR(p_2)\\
\forall p_1 \forall p_2 ~ML(p_1) \land Cite(p_1, p_2) \rightarrow ML(p_2)\\
\forall p_1 \forall p_2 ~HCI(p_1) \land Cite(p_1, p_2) \rightarrow HCI(p_2)
\end{array}
\end{equation*}
where $Cite(\cdot , \cdot)$ is an evidence predicate (e.g. its value over the groundings is known a-priori) determining whether a pattern cites another one. Different topics are differently closed with respect to other fields, and the above rules hold with different degrees.

A neural network with three hidden layers with $50$ units and RELU activation functions and one output layer using the softmax activation is used for this task as baseline. RNM employs the same network but with no output layer as the output layer is computed as part of the inference process as shown in Section~\ref{sec:dlm} for the one and multi-label classification cases. The Adam optimizer~\cite{kingma2014adam} is used to update the weights.
A variable portion of the data is sampled for training, of which $10$\% of this data is kept as validation set, while the remaining data is used as test set.
\begin{table}
\centering
\begin{tabular}{lccc}
\hline
\% training data & NN baseline & SBR & RNM\\
\hline
90               & 0.723       &  0.726 & 0.732\\
75               & 0.717       &  0.719 & 0.726\\
50               & 0.707       &  0.712 & 0.726\\
25               & 0.674       &  0.682 & 0.709\\
10               & 0.645       &  0.650 & 0.685\\
\hline
\end{tabular}
    \caption{Fully Observed Case. Results on the Citeseer dataset when using a subset of the supervised data for training using RNM, SBR and the baseline neural network.}
    \label{tab:citeseer}
\end{table}
\vspace{-0.3cm}
\paragraph{Fully Observed Case.}
The train and test sets are kept separated, and all links between train and test papers are dropped, so that the train and test data are two separate worlds.
Table~\ref{tab:citeseer} reports the result obtained by the baseline neural network, compared against the baseline model and SBR trained using the Lyrics framework as average over ten different samples of the train and test data. Since SBR can not learn the weight of the rules, these are validated by selecting the best performing one on the validation set. RNM improves over the other methods for all tested configurations, thanks to its ability of selecting the best weights for each rule, exploiting the fact that each research community has a different intra-community citation rate.

\begin{table}
\centering
\begin{tabular}{lccc}
\hline
\% training data & NN baseline & SBR & RNM\\
\hline
90               & 0.726 & 0.780 & 0.780\\
75               & 0.708 & 0.764 & 0.766\\
50               & 0.695 & 0.747 & 0.753\\
25               & 0.667 & 0.729 & 0.735\\
10               & 0.640 & 0.703 & 0.708\\
\hline
\end{tabular}
    \caption{Partially Observed Case. Results on the Citeseer dataset when using a subset of the supervised data for training using RNM, SBR and the baseline neural network.}
    \label{tab:citeseer_transductive}
\end{table}
\vspace{-0.3cm}
\paragraph{Partially Observed Case.} This experiment assumes that the training, validation and test data are available at training time~\cite{gammerman1998learning}, even if only the training labels are used during the training process. This configuration models a real world scenario where a partial knowledge of a world is given, but it is required to perform inference over the unknown portion of the environment. In this tranductive experiment, all Citeseer papers are supposed to be available in a single world together with the full citation network. Only a variable percentage of the supervised data is used during training. Therefore, the world is only partially observed at training time, and the EM schema described by Algorithm~\ref{al:training} must be used during training.

Table~\ref{tab:citeseer_transductive} reports the accuracy results obtained by the baseline neural network, compared against the baseline model and SBR. The SBR weights are validated by selecting the best performing one on the validation set. RNM improves over the other methods for all tested configurations, thanks to its ability of selecting the best weights for each rule.

\section{Conclusions and Future Work}
\label{sec:conclusions}
This paper presented Relational Neural Machines a novel framework to provide a tight integration between learning from supervised data and logic reasoning, allowing to improve the quality of both modules processing the low-level input data and the high-level reasoning about the environment. The presented model provides significant advantages over previous work in terms of scalability and flexiblity, while dropping any trade-off in exploiting the supervised data. The preliminary experimental results are promising, showing that the tighter integration between symbolic and a sub-symbolic levels helps in exploiting the input and output structures.
As future work, we plan to undertake a larger experimental exploration of RNM on real world problems for more structured problems.

\bibliography{references}

\begin{thebibliography}{10}

\bibitem{allamanis2017learning}
Miltiadis Allamanis, Pankajan Chanthirasegaran, Pushmeet Kohli, and Charles
  Sutton, `Learning continuous semantic representations of symbolic
  expressions', in {\em Proceedings of the 34th International Conference on
  Machine Learning-Volume 70}, pp. 80--88. JMLR. org, (2017).

\bibitem{bach2017hinge}
Stephen~H Bach, Matthias Broecheler, Bert Huang, and Lise Getoor, `Hinge-loss
  markov random fields and probabilistic soft logic', {\em Journal of Machine
  Learning Research}, {\bf 18},  1--67, (2017).

\bibitem{chen2015learning}
Liang-Chieh Chen, Alexander Schwing, Alan Yuille, and Raquel Urtasun, `Learning
  deep structured models', in {\em International Conference on Machine
  Learning}, pp. 1785--1794, (2015).

\bibitem{DeRaedtProbLog2007}
Luc De~Raedt, Angelika Kimmig, and Hannu Toivonen, `Problog: A probabilistic
  prolog and its application in link discovery', in {\em Proceedings of the
  20th International Joint Conference on Artifical Intelligence}, IJCAI'07, pp.
  2468--2473, San Francisco, CA, USA, (2007). Morgan Kaufmann Publishers Inc.

\bibitem{diligenti2017semantic}
Michelangelo Diligenti, Marco Gori, and Claudio Sacca, `Semantic-based
  regularization for learning and inference', {\em Artificial Intelligence},
  {\bf 244},  143--165, (2017).

\bibitem{donadello2017logic}
I~Donadello, L~Serafini, and AS~d'Avila Garcez, `Logic tensor networks for
  semantic image interpretation', in {\em IJCAI International Joint Conference
  on Artificial Intelligence}, pp. 1596--1602, (2017).

\bibitem{dong2018neural}
Honghua Dong, Jiayuan Mao, Tian Lin, Chong Wang, Lihong Li, and Denny Zhou,
  `Neural logic machines', in {\em International Conference on Learning
  Representations}, (2019).

\bibitem{dumanvcic2017demystifying}
Sebastijan Duman{\v{c}}i{\'c} and Hendrik Blockeel, `Demystifying relational
  latent representations', in {\em International Conference on Inductive Logic
  Programming}, pp. 63--77. Springer, (2017).

\bibitem{dumanvcic2019learning}
Sebastijan Duman{\v{c}}i{\'c}, Tias Guns, Wannes Meert, and Hendrik Blockeel,
  `Learning relational representations with auto-encoding logic programs', in
  {\em Proceedings of the 28th International Joint Conference on Artificial
  Intelligence}, pp. 6081--6087. AAAI Press, (2019).

\bibitem{gammerman1998learning}
Alexander Gammerman, Volodya Vovk, and Vladimir Vapnik, `Learning by
  transduction', in {\em Proceedings of the Fourteenth conference on
  Uncertainty in artificial intelligence}, pp. 148--155. Morgan Kaufmann
  Publishers Inc., (1998).

\bibitem{garcez2012neural}
Artur S~d'Avila Garcez, Krysia~B Broda, and Dov~M Gabbay, {\em Neural-symbolic
  learning systems: foundations and applications}, Springer Science \& Business
  Media, 2012.

\bibitem{giannini2018convex}
Francesco Giannini, Michelangelo Diligenti, Marco Gori, and Marco Maggini, `On
  a convex logic fragment for learning and reasoning', {\em IEEE Transactions
  on Fuzzy Systems}, (2018).

\bibitem{goodfellow2016deep}
Ian Goodfellow, Yoshua Bengio, Aaron Courville, and Yoshua Bengio, {\em Deep
  learning}, volume~1, MIT press Cambridge, 2016.

\bibitem{hajek2013metamathematics}
Petr H{\'a}jek, {\em Metamathematics of fuzzy logic}, volume~4, Springer
  Science \& Business Media, 2013.

\bibitem{hazan2016blending}
Tamir Hazan, Alexander~G Schwing, and Raquel Urtasun, `Blending learning and
  inference in conditional random fields', {\em The Journal of Machine Learning
  Research}, {\bf 17}(1),  8305--8329, (2016).

\bibitem{hu2016harnessing}
Zhiting Hu, Xuezhe Ma, Zhengzhong Liu, Eduard~H. Hovy, and Eric~P. Xing,
  `Harnessing deep neural networks with logic rules', in {\em Proceedings of
  the 54th Annual Meeting of the Association for Computational Linguistics,
  {ACL} 2016, August 7-12, 2016, Berlin, Germany, Volume 1: Long Papers},
  (2016).

\bibitem{jameel2016entity}
Shoaib Jameel and Steven Schockaert, `Entity embeddings with conceptual
  subspaces as a basis for plausible reasoning', in {\em Proceedings of the
  Twenty-second European Conference on Artificial Intelligence}, pp.
  1353--1361. IOS Press, (2016).

\bibitem{jiang2019neural}
Zhengyao Jiang and Shan Luo, `Neural logic reinforcement learning', {\em arXiv
  preprint arXiv:1904.10729}, (2019).

\bibitem{kathryn2018tensorlog}
William W Cohen Fan~Yang Kathryn and Rivard Mazaitis, `Tensorlog: Deep learning
  meets probabilistic databases', {\em Journal of Artificial Intelligence
  Research}, {\bf 1},  1--15, (2018).

\bibitem{kaur2017relational}
Navdeep Kaur, Gautam Kunapuli, Tushar Khot, Kristian Kersting, William Cohen,
  and Sriraam Natarajan, `Relational restricted boltzmann machines: A
  probabilistic logic learning approach', in {\em International Conference on
  Inductive Logic Programming}, pp. 94--111. Springer, (2017).

\bibitem{kimmig2012short}
Angelika Kimmig, Stephen Bach, Matthias Broecheler, Bert Huang, and Lise
  Getoor, `A short introduction to probabilistic soft logic', in {\em
  Proceedings of the NIPS Workshop on Probabilistic Programming: Foundations
  and Applications}, pp. 1--4, (2012).

\bibitem{kingma2014adam}
Diederik~P Kingma and Jimmy Ba, `Adam: A method for stochastic optimization',
  {\em arXiv preprint arXiv:1412.6980}, (2014).

\bibitem{Klement2000}
E.P. Klement, R.~Mesiar, and E.~Pap, {\em Triangular Norms}, Kluwer Academic
  Publisher, 2000.

\bibitem{lecun2015deep}
Yann LeCun, Yoshua Bengio, and Geoffrey Hinton, `Deep learning', {\em nature},
  {\bf 521}(7553),  436, (2015).

\bibitem{lippi2009betaresidue}
Marco Lippi and Paolo Frasconi, `Prediction of protein $\beta$-residue contacts
  by markov logic networks with grounding--specific weights', {\em
  Bioinformatics}, {\bf 25}(18),  2326--2333, (2009).

\bibitem{lu2003link}
Qing Lu and Lise Getoor, `Link-based classification', in {\em Proceedings of
  the 20th International Conference on Machine Learning (ICML-03)}, pp.
  496--503, (2003).

\bibitem{manhaeve2018deepproblog}
Robin Manhaeve, Sebastijan Duman{\v{c}}i{\'c}, Angelika Kimmig, Thomas
  Demeester, and Luc De~Raedt, `Deepproblog: Neural probabilistic logic
  programming', {\em arXiv preprint arXiv:1805.10872}, (2018).

\bibitem{marra2019integrating}
Giuseppe Marra, Francesco Giannini, Michelangelo Diligenti, and Marco Gori,
  `Integrating learning and reasoning with deep logic models', in {\em
  Proceedings of the European Conference on Machine Learning}, (2019).

\bibitem{marra2019lyrics}
Giuseppe Marra, Francesco Giannini, Michelangelo Diligenti, and Marco Gori,
  `Lyrics: a general interface layer to integrate ai and deep learning', {\em
  arXiv preprint arXiv:1903.07534}, (2019).

\bibitem{marra2019neural}
Giuseppe Marra and Ond{\v{r}}ej Ku{\v{z}}elka, `Neural markov logic networks',
  {\em arXiv preprint arXiv:1905.13462}, (2019).

\bibitem{minervini2017regularizing}
Pasquale Minervini, Luca Costabello, Emir Mu{\~n}oz, V{\'\i}t
  Nov{\'a}{\v{c}}ek, and Pierre-Yves Vandenbussche, `Regularizing knowledge
  graph embeddings via equivalence and inversion axioms', in {\em Joint
  European Conference on Machine Learning and Knowledge Discovery in
  Databases}, pp. 668--683. Springer, (2017).

\bibitem{muggleton1994inductive}
Stephen Muggleton and Luc De~Raedt, `Inductive logic programming: Theory and
  methods', {\em The Journal of Logic Programming}, {\bf 19},  629--679,
  (1994).

\bibitem{nickel2016holographic}
Maximilian Nickel, Lorenzo Rosasco, and Tomaso Poggio, `Holographic embeddings
  of knowledge graphs', in {\em Thirtieth Aaai conference on artificial
  intelligence}, (2016).

\bibitem{niepert2016discriminative}
Mathias Niepert, `Discriminative gaifman models', in {\em Advances in Neural
  Information Processing Systems}, pp. 3405--3413, (2016).

\bibitem{richardson2006markov}
Matthew Richardson and Pedro Domingos, `Markov logic networks', {\em Machine
  learning}, {\bf 62}(1),  107--136, (2006).

\bibitem{rocktaschel2016learning}
Tim Rockt{\"a}schel and Sebastian Riedel, `Learning knowledge base inference
  with neural theorem provers', in {\em Proceedings of the 5th Workshop on
  Automated Knowledge Base Construction}, pp. 45--50, (2016).

\bibitem{rocktaschel2017end}
Tim Rockt{\"a}schel and Sebastian Riedel, `End-to-end differentiable proving',
  in {\em Advances in Neural Information Processing Systems}, pp. 3788--3800,
  (2017).

\bibitem{santoro2017simple}
Adam Santoro, David Raposo, David~G Barrett, Mateusz Malinowski, Razvan
  Pascanu, Peter Battaglia, and Timothy Lillicrap, `A simple neural network
  module for relational reasoning', in {\em Advances in neural information
  processing systems}, pp. 4967--4976, (2017).

\bibitem{serafini2017learning}
Luciano Serafini, Ivan Donadello, and Artur~d'Avila Garcez, `Learning and
  reasoning in logic tensor networks: theory and application to semantic image
  interpretation', in {\em Proceedings of the Symposium on Applied Computing},
  pp. 125--130. ACM, (2017).

\bibitem{sourek2018lifted}
Gustav Sourek, Vojtech Aschenbrenner, Filip Zelezny, Steven Schockaert, and
  Ondrej Kuzelka, `Lifted relational neural networks: Efficient learning of
  latent relational structures', {\em Journal of Artificial Intelligence
  Research}, {\bf 62},  69--100, (2018).

\bibitem{sutton2007piecewise}
Charles Sutton and Andrew McCallum, `Piecewise pseudolikelihood for efficient
  training of conditional random fields', in {\em Proceedings of the 24th
  international conference on Machine learning}, pp. 863--870. ACM, (2007).

\bibitem{sutton2012introduction}
Charles Sutton, Andrew McCallum, et~al., `An introduction to conditional random
  fields', {\em Foundations and Trends{\textregistered} in Machine Learning},
  {\bf 4}(4),  267--373, (2012).

\bibitem{wang2015knowledge}
Quan Wang, Bin Wang, and Li~Guo, `Knowledge base completion using embeddings
  and rules', in {\em Twenty-Fourth International Joint Conference on
  Artificial Intelligence}, (2015).

\bibitem{xu2017semantic}
Jingyi Xu, Zilu Zhang, Tal Friedman, Yitao Liang, and Guy Van~den Broeck, `A
  semantic loss function for deep learning with symbolic knowledge', in {\em
  Proceedings of the 35th International Conference on Machine Learning (ICML)},
  (July 2018).

\bibitem{yang2017differentiable}
Fan Yang, Zhilin Yang, and William~W Cohen, `Differentiable learning of logical
  rules for knowledge base reasoning', in {\em Advances in Neural Information
  Processing Systems}, pp. 2319--2328, (2017).

\end{thebibliography}
\end{document}